\documentclass[11pt,a4paper]{article}
\usepackage[hyperref]{acl2018}
\usepackage{times}
\usepackage{latexsym}

\usepackage[T1]{fontenc}
\usepackage{dsfont}

\usepackage{mdframed}
\usepackage{booktabs}
\usepackage{graphicx}

\usepackage[normalem]{ulem}

\usepackage{xcolor}
\usepackage{soul}
\colorlet{soulred}{red!30}
\sethlcolor{soulred}%

\usepackage{amsmath,amsthm,amsfonts,amssymb,bm}

\DeclareMathOperator*{\argmax}{arg\,max} % thin space, limits on side in displays
\usepackage{framed}
\usepackage{varioref}
\usepackage{float}
\usepackage{multirow}
\usepackage{dashrule}
\usepackage{url}

\aclfinalcopy % Uncomment this line for the final submission
\def\aclpaperid{1003} %  Enter the acl Paper ID here

\title{Harvesting Paragraph-Level Question-Answer Pairs \\ 
        from Wikipedia}

\author{Xinya Du \ {\normalfont and} \ Claire Cardie\\
  Department of Computer Science\\
  Cornell University \\
  Ithaca, NY, 14853, USA \\
  {\tt \{xdu, cardie\}@cs.cornell.edu} \\
  }

\date{}

\begin{document}
\maketitle
\begin{abstract}
We study the task of generating from Wikipedia articles question-answer pairs that cover content beyond a single sentence. We propose a neural network approach that incorporates  coreference knowledge via a novel gating mechanism. Compared to models that only take into account sentence-level information~\cite{heilman2010good, du2017LearningToAsk, zhou2017neural}, we find that the linguistic knowledge introduced by the coreference representation aids question generation significantly, producing models that outperform the current state-of-the-art. We apply our system (composed of an answer span extraction system and the passage-level QG system) to the 10,000 top-ranking Wikipedia articles and create a corpus of over one million question-answer pairs. We also provide a qualitative analysis for this large-scale generated corpus from Wikipedia. %and post-filter the pairs with a reading comprehension model. 
\end{abstract}

\section{Introduction}

Recently, there has been a resurgence of work in NLP on reading comprehension~\cite{hermann2015teaching, rajpurkar2016squad, JoshiTriviaQA2017} with the goal of developing systems that can answer questions about the content of a given passage or document. Large-scale QA datasets are indispensable for training expressive statistical models for this task and play a critical role in advancing the field. And there have been a number of efforts in this direction.~\newcite{miller2016key}, for example, develop a dataset for open-domain
question answering;~\newcite{rajpurkar2016squad} and \newcite{JoshiTriviaQA2017} do so for reading comprehension (RC); and ~\newcite{hill2015goldilocks} and \newcite{hermann2015teaching}, for the related task of answering cloze questions~\cite{winograd1972understanding, levesque2011winograd}. To create these datasets, either crowdsourcing or {\em (semi-)synthetic} approaches are used. The (semi-)synthetic datasets (e.g., \newcite{hermann2015teaching}) are large in size and cheap to obtain; however, they do not share the same characteristics as explicit  QA/RC questions~\cite{rajpurkar2016squad}. In comparison, high-quality {\em crowdsourced} datasets are much smaller in size, and the annotation process is quite expensive because the labeled examples require expertise and careful design~\cite{danqi2016exam}.

\begin{figure}[t]
\centering
% \toprule[0.05cm]
\begin{tabular}{p{6.5cm}} 
\toprule
    \small \textbf{Paragraph}:\\
    {
    \fontfamily{phv} \selectfont 
    \small \textsuperscript{(1)}\textit{Tesla} was renowned for \textit{his} achievements and showmanship, eventually earning \textit{him} a reputation in popular culture as an archetypal "\colorlet{soulred}{red!30}\hl{mad scientist}". \textsuperscript{(2)}\textit{His} \colorlet{soulred}{green!40}\hl{patents} earned \textit{him} a considerable amount of money, much of which was used to finance \textit{his} own projects with varying degrees of success. \textsuperscript{(3)}\textit{He} lived most of his life in a series of \colorlet{soulred}{blue!40}\hl{New York hotels}, through \textit{his} retirement. \textsuperscript{(4)}\textit{Tesla} died on 7 January 1943. ...
    }\\[4pt]
    
    \small \textbf{Questions}: \\
    
    % 1st question
    {
    \small
    \fontfamily{phv}\selectfont 
    --  What was Tesla's reputation in popular culture?
    }\\
     {
    \small
    \fontfamily{phv}\selectfont 
    \quad \emph{\textcolor{red}{mad scientist}}
    }\\[4pt]
    
    % 2st question
    {
    \small
    \fontfamily{phv}\selectfont 
    -- How did Tesla finance his work?
    }\\
    {
    \small
    \fontfamily{phv}\selectfont 
    \quad \emph{\textcolor{green!100}{patents}}
    }\\[4pt]
    
    % 2nd question
    {
    \small \fontfamily{phv}\selectfont 
    -- Where did Tesla live for much of his life?
    }\\
    {
    \small \fontfamily{phv}\selectfont 
    \quad \emph{\textcolor{blue}{New York hotels}}
    }\\[4pt]

\bottomrule
\end{tabular}
% \bottomrule[0.05cm]
\vspace{-0.2cm}
\caption{Example input from the fourth paragraph of a Wikipedia article on ~\textit{Nikola Tesla}, along with the natural questions and their answers from the SQuAD~\cite{rajpurkar2016squad} dataset. We show in italics the set of mentions that refer to Nikola Tesla --- {\em Tesla}, {\em him}, {\em his}, {\em he}, etc.}
% The mentions in the cluster \{Tesla, his, him\} (italicized text in example) all refer to the same entity -- Nikola Tesla. }
 \label{fig:problem}
\vspace{-0.5cm}
\end{figure}

Thus, there is a need for methods that can automatically generate high-quality question-answer pairs.~\newcite{serban2016factoid30m} propose the use of recurrent neural networks to generate QA pairs from structured knowledge resources such as Freebase. Their work relies on the existence of  automatically acquired KBs, which are known to have errors and suffer from incompleteness. They are also non-trivial to obtain. In addition, the questions in the resulting dataset are limited  to queries regarding a single fact (i.e., tuple) in the KB.

% However, their work was based on SimpleQuestions~\cite{bordes2015large}, which is a single-relation question-answering dataset on Freebase triples (subject, relationship, object), the question format is always about a single fact of the knowledge base (KB) and is very constrained. Besides, \newcite{serban2016factoid30m}'s approach relies on KB which has inherent limitations (incompleteness, fixed schemas).

Motivated by the need for large scale QA pairs and the limitations of recent work, we investigate methods that can automatically ``harvest'' (generate) question-answer pairs from raw text/unstructured documents, such as Wikipedia-type articles.

Recent work along these lines~\cite{du2017LearningToAsk, zhou2017neural} (see Section~\ref{sec:related}) has proposed the use of attention-based recurrent neural models trained on the crowdsourced SQuAD dataset~\cite{rajpurkar2016squad} for question generation. While successful, the resulting QA pairs are based on information from a single sentence. As described in~\newcite{du2017LearningToAsk}, however, nearly 30\% of the questions in the human-generated questions of SQuAD rely on information beyond a single sentence. For example, in Figure~\ref{fig:problem}, the second and third questions require coreference information (i.e., recognizing that ``His'' in sentence 2 and ``He'' in sentence 3 both corefer with ``Tesla'' in sentence 1) to answer them.

% Motivated by the improvement for document-level understanding (e.g., dialogue systems~\cite{strube2003machine, muller2007resolving, seo2017visual}, summarization~\cite{stoyanov2006partially}.) brought by coreference resolution, whose own performance has also witnessed huge gain with application of deep representation learning~\cite{clark2016improving, lee2017end}.
Thus, our research studies methods for incorporating coreference information into the training of a question generation system. In particular, we propose gated {\bf Coref}erence knowledge for {\bf N}eural {\bf Q}uestion {\bf G}eneration ({\bf CorefNQG}), a neural sequence model with a novel gating mechanism that leverages continuous representations of {\em coreference clusters} --- the set of mentions used to refer to each entity --- to better encode linguistic knowledge introduced by coreference, for paragraph-level question generation. 

In an evaluation using the SQuAD dataset, we find that CorefNQG 
%encodes coreference knowledge in training the neural sequence model for 
enables better question generation. It outperforms significantly the baseline neural sequence models that encode information from a single sentence, and a model that encodes \textit{all} preceding context and the input sentence itself. When evaluated on only the portion of SQuAD that requires coreference resolution, the gap between our system and the baseline systems is even larger.

By applying our approach to the 10,000 top-ranking Wikipedia articles, we obtain a question answering/reading comprehension dataset with over one million QA pairs; we provide a qualitative analysis in Section~\ref{sec:res}. The dataset and the source code for the system are available at \url{https://github.com/xinyadu/HarvestingQA}.

% We release the the first large scale QA corpus automatically generated corpus from Wikipedia.

% In the following sections, we describe related work (Section 2);  ... (Section 7) and conclusion (Section 8).

% the most probable coreference resolution mention after the original mention, to introduce external information beyond sentence-level~\cite{clark2016improving}.

% Currently coreference, entailment, parsing, and other NLP 

\section{Related Work}
\label{sec:related}

\subsection{Question Generation}

Since the work by~\newcite{rus2010first}, question generation (QG) has attracted interest from both the NLP and NLG communities. Most early work in QG employed rule-based approaches to transform input text into questions, usually requiring the application of a sequence of well-designed general rules or templates~\cite{mitkov2003computer, labutov2015deep}.~\newcite{heilman2010good} introduced an overgenerate-and-rank approach: their system generates a set of questions and then ranks them to select the top candidates. Apart from generating questions from raw text, there has also been research on question generation from symbolic representations~\cite{yao2012semantics, olney2012question}. 
% For example,~\newcite{yao2012semantics}'s approach first parses the input sentence into its corresponding Minimal Recursion Semantics (MRS) representation and then generate a question with lexicon and grammar rules.

With the recent development of deep representation learning and large QA datasets, there has been research on recurrent neural network based approaches for question generation.~\newcite{serban2016factoid30m} used the encoder-decoder framework to generate QA pairs from knowledge base triples; ~\newcite{mitesh2017generating} generated questions from a knowledge graph; ~\newcite{du2017LearningToAsk} studied how to generate questions from sentences using an attention-based sequence-to-sequence model and investigated the effect of exploiting sentence- vs.\ paragraph-level information.~\newcite{du2017identifying} proposed a hierarchical neural sentence-level sequence tagging model for identifying question-worthy sentences in a text  passage. Finally, \newcite{duan2017qgforqa} investigated how to use question generation to help improve question answering systems on the sentence selection subtask. 

In comparison to the related methods from above that generate questions from raw text, our method is different in its ability to take into account contextual information beyond the sentence-level by introducing coreference knowledge.
% from images~\cite{nasrin2016vqg}, exploring the deeper connection between vision and language.

\subsection{Question Answering Datasets and Creation}

Recently there has been an increasing interest in question answering with the creation of many datasets. Most are built using crowdsourcing; they are generally comprised of fewer than 100,000 QA pairs and are time-consuming to create. WebQuestions~\cite{berant2013semantic}, for example, contains 5,810 questions crawled via the Google Suggest API and is designed for knowledge base QA with answers restricted to Freebase entities. To tackle the size issues associated with  WebQuestions,~\newcite{bordes2015large} introduce SimpleQuestions, a dataset of 108,442 questions authored by English speakers. SQuAD~\cite{rajpurkar2016squad} is a dataset for machine comprehension; it is created by showing a Wikipedia paragraph to human annotators and asking them to write questions based on the paragraph. TriviaQA~\cite{JoshiTriviaQA2017} includes 95k question-answer authored by trivia enthusiasts and corresponding evidence documents.
% Quasar-T~\cite{dhingra2017quasar}
% Reading Wikipedia to answer open-domain questions~\cite{chen2017reading}

(Semi-)synthetic generated datasets are easier to build to large-scale~\cite{hill2015goldilocks, hermann2015teaching}. They usually come in the form of cloze-style questions. For example,~\newcite{hermann2015teaching} created over a million examples by pairing CNN and Daily Mail news articles with their summarized bullet points.~\newcite{danqi2016exam} showed that this dataset is quite noisy due to the method of data creation and concluded that performance of QA systems on the dataset is almost saturated.

Closest to our work is that of~\newcite{serban2016factoid30m}. They train a neural triple-to-sequence model on SimpleQuestions, and apply their system to Freebase to produce a large collection of human-like question-answer pairs.

% There has also been work on automatically creating question-answer pairs in the information retrieval community, but they mainly focus on the setting of community question answering.~\newcite{cong2008finding} worked on online forum question detection and QA extraction.

% \textbf{linguistics for deep nlp, coref}
% Modeling Source Syntax for Neural Machine Translation~\cite{modeling2017li}
% End-to-end Neural Coreference Resolution~\cite{lee2017end}
% gating mechanism~\cite{van2016pixel, dauphin2017language}

\section{Task Definition}
%\subsection{Candidate Answer Extraction}
\label{sec:ans}
Our goal is to harvest high quality question-answer pairs from the paragraphs of an article of interest. In our task formulation, 
this consists of two steps: {\bf candidate answer extraction} and 
{\bf answer-specific question generation}.  
Given an input paragraph, we % of $n$ sentences, we % $x = (x_1, x_2, ..., x_n)$, we
first identify a set of {\em question-worthy} candidate answers  
$ans = (ans_1, ans_2, ..., ans_l)$, each a span of text as denoted in color in
Figure~\ref{fig:problem}.
%The first step of harvesting question-answer pairs from the articles is to identify a set of {\em question-worthy} candidate answer spans (e.g., the colored spans in Figure~\ref{fig:problem}). Given an input paragraph $x = (x_1, x_2, ..., x_n)$, where $n$ is the length of the paragraph and $x_i$ is the word in the sentence. The extraction system produces $a = (a_1, a_2, ..., a_l)$, where $l$ is the number of answer spans.
%
%\subsection{Transducing Answer-specific Sentences to Questions}
For each candidate answer $ans_i$, we then aim to generate a question $Q$ ---
a sequence of tokens $y_1,..., y_N$ ---
based on the sentence $S$ that contains candidate $ans_i$ such that: 
% based on the sentence that contains it such that:
\begin{itemize}
\itemsep-0.15em 
  \item $Q$ asks about an aspect of $ans_i$ that is of potential 
  interest to a human; % question asker.
  %\item $a_i$ represents a valid answer to the question.
  \item $Q$ might rely on information from sentences that precede $S$ in 
  the paragraph.
\end{itemize}
Mathematically then, 
\begin{equation}
\label{equ:task}
% \mathbf{\overline{y}} = \argmax_{\mathbf{y}} P \left( \mathbf{y} \vert \mathbf{x} \right)
Q = \argmax_{Q} P \left( Q \vert S, C \right)
\end{equation}
where  
$P(Q|S,C) = \prod_{n=1}^{N} P \left(y_n \vert y_{<n}, S,C \right)$ 
where $C$ is the set of sentences that precede $S$ in the paragraph.

\section{Methodology}

\begin{figure*}[t]
\centering
\small
\includegraphics[scale=.35]{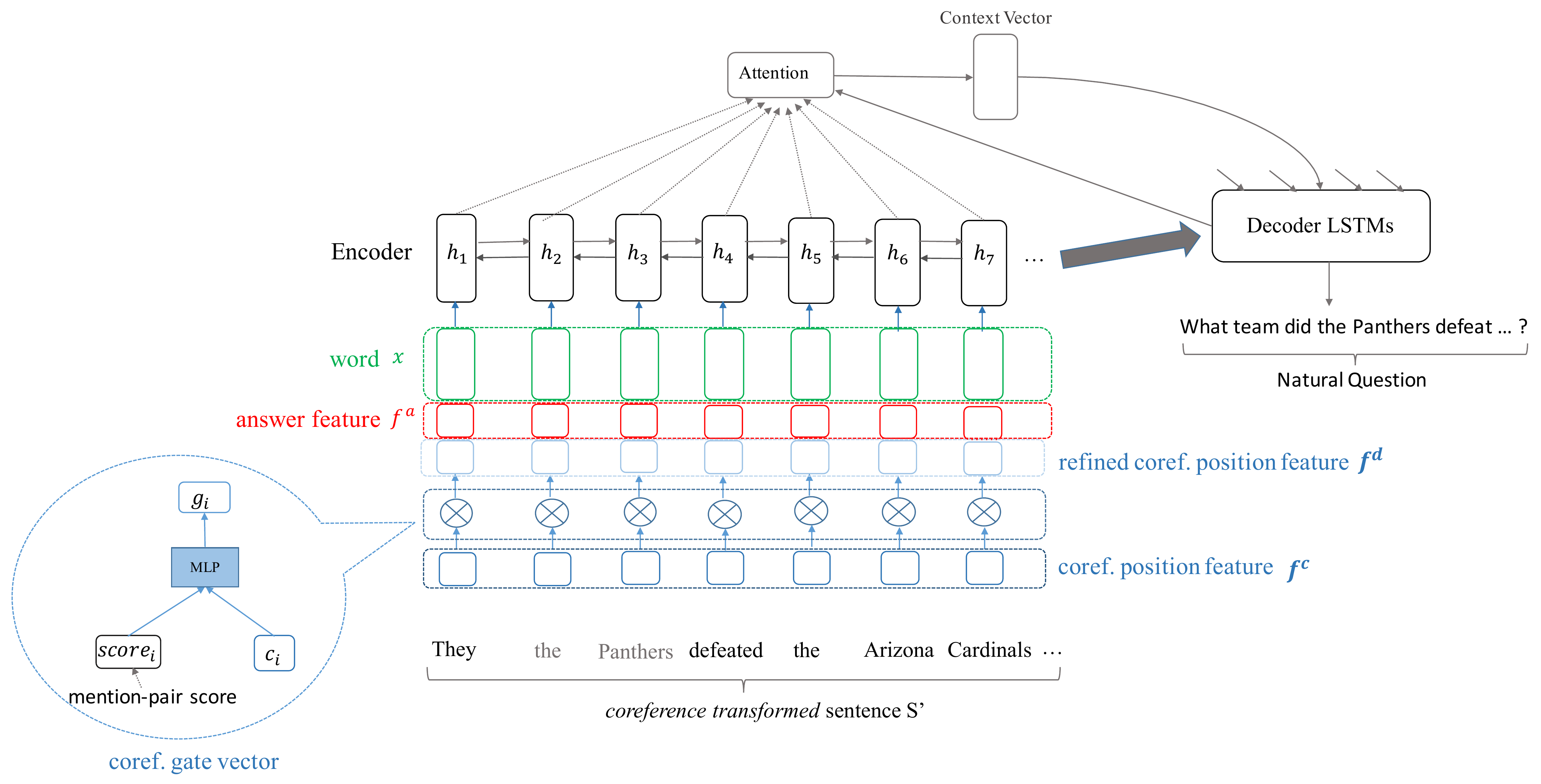}
\vspace{-0.5cm}
\caption{The gated {\bf Coref}erence knowledge for {\bf N}eural {\bf Q}uestion {\bf G}eneration ({\bf CorefNQG}) Model.}
\label{fig:nn}
\vspace{-0.1cm}
\end{figure*}

In this section, we introduce our framework for harvesting the question-answer pairs.
As described above, it consists of the question generator CorefNQG
(Figure~\ref{fig:nn}) and a %NER-feature enhanced 
candidate answer extraction module. During test/generation time, we 
(1) run the answer extraction module on the input text to obtain answers, 
and then (2) run the question generation module to obtain the corresponding questions.

% fea_example
\begin{table*}[!htb]
\begin{tabular}{ l|c c c c c c c c c c c l}
\toprule
{ word} & they & the  & panthers & defeated &  the & arizona & cardinals & 49  & -- & 15 & ... \\ 
{ans. feature} & \verb|O| & \verb|O| & \verb|O| & \verb|O| & \verb|B_ANS|& \verb|I_ANS| & \verb|I_ANS| & \verb|O| & \verb|O| & \verb|O| & ... \\ 
{coref. feature} & \verb|B_PRO| & \verb|B_ANT| & \verb|I_ANT| & \verb|O| &  \verb|O| & \verb|O| & \verb|O| & \verb|O| & \verb|O| & \verb|O| & ...\\  
  \bottomrule
\end{tabular}
\vspace{-0.2cm}
\caption{Example input sentence with coreference and answer position features. 
%``they'' is the pronoun and ``the panthers'' is the antecedent. 
The corresponding gold question is ``What team did the Panthers defeat in the NFC championship game ?''}
\vspace{-0.5cm}
\label{tab:fea_example}
\end{table*}

\subsection{Question Generation}

As shown in Figure~\ref{fig:nn}, our generator prepares the feature-rich input embedding --- a concatenation of (a) a refined coreference position feature embedding, (b) an answer feature embedding, and (c) a word embedding, each of which is described below.  It then encodes the textual input using an LSTM unit~\cite{hochreiter1997long}. Finally, an attention-copy equipped decoder is used to decode the question. 

More specifically, given the input sentence $S$ (containing an answer span) and the preceding context $C$, we first run a coreference resolution system to get the coref-clusters for $S$ and $C$ and use
them to create a {\em coreference transformed} input sentence: for each pronoun, we append its most representative non-pronominal coreferent mention. Specifically, we apply the simple feedforward network based mention-ranking model of ~\newcite{clark2016improving} to the concatenation of $C$ and $S$ to get the coref-clusters for all entities in $C$ and $S$. The C\&M model produces a score/representation $s$ for each mention pair $(m_1,m_2)$,
% , and each mention is included in the cluster with the antecedent that it has the highest score with,
\vspace{-0.1cm}
\begin{equation}
\label{equ:score}
% \vspace{-0.3cm}
    s(m_1, m_2) = \mathbf{W}_m h_m(m_1, m_2) + b_m
\end{equation}
where $\mathbf{W}_m$ is a $1 \times d$ weight matrix and b is the bias. $h_m(m_1,m_2)$ is representation of the last hidden layer of the three layer feedforward neural network.

For each pronoun in $S$, we then heuristically identify the most ``representative'' antecedent from its coref-cluster. (Proper nouns are preferred.)   We append the new mention after the pronoun. For example, in Table~\ref{tab:fea_example}, ``the panthers'' is the most representative mention in the coref-cluster for ``they''. The new sentence with the appended coreferent mention is our {\em coreference transformed} input sentence $S^{'}$ (see Figure~\ref{fig:nn}).

\vspace{0.2cm} \noindent {\bf Coreference Position Feature Embedding} \quad 
For each token in $S^{'}$, we also maintain one position feature $\mathbf{f^c} = (c_1,...,c_n)$, to denote pronouns (e.g., ``they'') and antecedents (e.g., ``the panthers''). We use the BIO tagging scheme to label the associated spans in $S^{'}$. ``\verb|B_ANT|'' denotes the start of an antecedent span, tag ``\verb|I_ANT|'' continues the antecedent span and tag ``\verb|O|'' marks tokens that do not form part of a mention span. Similarly, tags ``\verb|B_PRO|'' and ``\verb|I_PRO|'' denote the pronoun span. (See Table~\ref{tab:fea_example}, ``coref.\ feature''.)

\vspace{0.2cm} \noindent {\bf Refined Coref.\ Position Feature Embedding} \quad Inspired by the success of gating mechanisms for controlling information flow in neural networks~\cite{hochreiter1997long, dauphin2017language}, we propose to use a gating network here to obtain a refined representation of the coreference position feature vectors  $\mathbf{f^c} = (c_1,...,c_n)$. The main idea is to utilize the mention-pair score (see Equation~\ref{equ:score}) to help the neural network learn the importance of the coreferent phrases. We compute the refined (gated) coreference position feature vector $\mathbf{f^d} = (d_1,...,d_n)$ as follows,
\begin{equation}
\begin{gathered}
\label{equ:gate}
% \mathbf{\overline{y}} = \argmax_{\mathbf{y}} P \left( \mathbf{y} \vert \mathbf{x}
g_i = \textnormal{ReLU}(\mathbf{W}_a c_i + \mathbf{W}_b score_i + b)\\
d_i = g_i \odot c_i
\vspace{-0.1cm}
\end{gathered}
\end{equation}
where $\odot$ denotes an element-wise product between two vectors and ReLU is the rectified linear activation function. $score_i$ denotes the mention-pair score for each antecedent token (e.g., ``the'' and ``panthers'') with the pronoun (e.g., ``they''); $score_i$ is obtained from the trained model (Equation~\ref{equ:score}) of the C\&M. If token $i$ is not added later as an antecedent token, $score_i$ is set to zero. $\mathbf{W}_a$, $\mathbf{W}_b$ are weight matrices and $b$ is the bias vector.

\vspace{0.2cm} \noindent {\bf Answer Feature Embedding} \quad We also include an answer position feature embedding to generate answer-specific questions; we denote the answer span with the usual BIO tagging scheme (see, e.g., ``the arizona cardinals'' in Table~\ref{tab:fea_example}).  During training and testing, the answer span feature (i.e., ``\verb|B_ANS|'', ``\verb|I_ANS|'' or ``\verb|O|'') is mapped to its feature embedding space: $\mathbf{f^a} = (a_1, ..., a_n)$.

\vspace{0.2cm} \noindent {\bf Word Embedding} \quad To obtain the word embedding for the tokens themselves, we just map the tokens to the word embedding space: $\mathbf{x} = (x_1,...,x_n)$.

\vspace{0.25cm} \noindent {\bf Final Encoder Input} \quad As noted above, the final input to the LSTM-based encoder is a concatenation of (1) the refined coreference position feature embedding (light blue units in Figure~\ref{fig:nn}), (2) the answer position feature embedding (red units), and (3) the word embedding for the token (green units),
\begin{equation}
e_i = \textnormal{concat}(d_i, a_i, x_i)
\vspace{-0.1cm}
\end{equation}

\vspace{0.25cm} \noindent {\bf Encoder} \quad As for the encoder itself, we use bidirectional LSTMs to read the input $\mathbf{e} = (e_1,...,e_n)$ in both the forward and backward directions. After encoding, we obtain two sequences of hidden vectors, namely, $\overrightarrow{\mathbf{h}} = (\overrightarrow{h_1},..., \overrightarrow{h_n})$ and $\overleftarrow{\mathbf{h}} = (\overleftarrow{h_1},..., \overleftarrow{h_n})$. The final output state of the encoder is the concatenation of $\overrightarrow{\mathbf{h}}$ and $\overleftarrow{\mathbf{h}}$ where 
\begin{equation}
h_i = \textnormal{concat}(\overrightarrow{h_i}, \overleftarrow{h_i})
\vspace{-0.1cm}
\end{equation}

\vspace{0.25cm} \noindent {\bf Question Decoder with Attention \& Copy} On top of the feature-rich encoder, we use LSTMs with attention~\cite{bahdanau2014neural} as the decoder for generating the question $y_1,..., y_m$ one token at a time. To deal with rare/unknown words, the decoder also allows directly copying words from the source sentence via pointing~\cite{vinyals2015pointer}.

At each time step $t$, the decoder LSTM reads the previous word embedding $w_{t-1}$ and previous hidden state $s_{t-1}$ to compute the new hidden state, 
    \vspace{-0.1cm}
\begin{equation}
    s_t = \textnormal{LSTM} (w_{t-1}, s_{t-1})
    \vspace{-0.1cm}
\end{equation}
\noindent
Then we calculate the {\em attention distribution} $\alpha_t$ as in~\newcite{bahdanau2014neural}, 
\begin{equation}
\begin{gathered}
    e_{t,i} = h_i^{T}\mathbf{W}_c s_{t-1} \\
    \alpha_{t} = \textnormal{softmax}(e_{t})
\end{gathered}
\end{equation}
where $\mathbf{W}_c$ is a weight matrix and attention distribution $\alpha_t$ is a probability distribution over the source sentence words. With $\alpha_t$, we can obtain the context vector $h^{*}_t$,
\begin{equation}
    h^{*}_t = \sum_{i=1}^{n} \alpha_t^{i}h_i
    \vspace{-0.1cm}
\end{equation}
\noindent
Then, using the context vector $h_t^{*}$ and hidden state $s_t$, the probability distribution over the target (question) side vocabulary is calculated as,
\begin{equation}
    P_{vocab} = \textnormal{softmax}(\mathbf{W}_d \textnormal{concat}(h_t^{*}, s_t))
\end{equation}
\noindent
Instead of directly using $P_{vocab}$ for training/generating with the fixed target side vocabulary, we also consider {\em copying} from the source sentence. The copy probability is based on the context vector $h_t^{*}$ and hidden state $s_t$, 
\begin{equation}
    \lambda_{t}^{copy} = \sigma \left( \mathbf{W}_e h^{*}_t + \mathbf{W}_f s_t\right)
    \vspace{-0.1cm}
\end{equation}
\noindent
and the probability distribution over the source sentence words is the sum of the attention scores of the corresponding words, 
% $P_{copy}(w) = \sum_{i=1}^{n} \alpha_t^{i} * \mathds{1}\{w == w_i\}$
\begin{equation}
P_{copy}(w) = \sum_{i=1}^{n} \alpha_t^{i} * \mathds{1}\{w == w_i\}
\vspace{-0.2cm}
\end{equation}
\noindent
Finally, we obtain the probability distribution over the dynamic vocabulary (i.e., union of original target side and source sentence vocabulary) by summing over $P_{copy}$ and $P_{vocab}$,
\begin{equation}
    P(w) =  \lambda_t^{copy} P_{copy}(w) + (1 -\lambda_t^{copy}) P_{vocab}(w)
    % \vspace{-0.2cm}
\end{equation}
where $\sigma$ is the sigmoid function, and $\mathbf{W}_d$, $\mathbf{W}_e$, $\mathbf{W}_f$ are weight matrices.

% \vspace{0.25cm} \noindent {\bf Copy Mechanism} ~\newcite{gulcehre2016pointing} propose to use pointing mechanism to copy rare words from source sentence, to deal with the rare and unknown words problem. In question generation, there might also be rare words in the source sentence that is out of the vocabulary. We employ the mechanism during our decoding process.
\subsection{Answer Span Identification}
We frame the problem of identifying candidate answer spans from a paragraph as a sequence labeling task and base our model on the BiLSTM-CRF approach for named entity recognition~\cite{huang2015bidirectional}. 
% As mentioned in~\cite{rajpurkar2016squad}, answers in the SQuAD dataset can be categorized into ten types, i.e., ``date'', ``other numeric'', ``person'', ``location'', ``other entity'', ``common noun phrase'', ``adjective phrase'', ``verb phrase'', ``clause'' and ``other'', 
Given a paragraph of $n$ tokens, instead of directly feeding the sequence of word vectors $\mathbf{x} = (x_1,...,x_n)$ to the LSTM units, we first construct the feature-rich embedding $\mathbf{x}^{'}$ for each token, which is the concatenation of the word embedding, an NER feature embedding, and a character-level representation of the word~\cite{lample2016neural}. We use the concatenated vector as the ``final'' embedding $\mathbf{x}^{'}$ for the token, 
\begin{equation}
    x_{i}^{'} = \textnormal{concat}(x_i, \textnormal{CharRep}_i, \textnormal{NER}_i) 
\end{equation}
where $\textnormal{CharRep}_i$ is the concatenation of the last hidden states of a character-based biLSTM. %forward and backward sequences running on the characters of the word. 
The intuition behind the use of NER features is that SQuAD answer spans contain a large number of named entities,
% common noun phrases, 
numeric phrases, etc.

Then a multi-layer Bi-directional LSTM is applied to $(x_1^{'},..., x_n^{'})$ and we obtain the output state $z_{t}$ for time step $t$ by concatenation of the hidden states (forward and backward) at time step $t$ from the last layer of the BiLSTM. We apply the $\textnormal{softmax}$ to $(z_1, ..., z_n)$ to get the normalized score representation for each token, which is of size $n \times k$, where $k$ is the number of tags.  

Instead of using a softmax training objective that minimizes the cross-entropy loss for each individual word, the model is trained with a CRF~\cite{lafferty2001conditional} objective, which minimizes the negative log-likelihood for the entire correct sequence: $-\log(p_\mathbf{y})$, 
\begin{equation}
    p_\mathbf{y} = \frac{\textnormal{exp}(q(\mathbf{x}^{'},\mathbf{y}))}{\sum_{\mathbf{y}^{'} \in \mathbf{Y}^{'}}\textnormal{exp}(q(\mathbf{x}^{'},\mathbf{y}^{'}))}
\end{equation}
where $q(\mathbf{x}^{'},\mathbf{y}) = \sum_{t=1}^{n} P_{t,y_t} + \sum_{t=0}^{n-1}A_{y_t,y_{t+1}}$, $P_{t,y_t}$ is the score of assigning tag $y_t$ to the $t^{th}$ token, and $A_{y_t,y_{t+1}}$ is the transition score from tag $y_t$ to $y_{t+1}$, the scoring matrix $A$ is to be learned. $\mathbf{Y}^{'}$ represents all the possible tagging sequences.
% Finally, we add a CRF layer~\cite{lafferty2001conditional} to jointly score the labeling decisions. For a sequence of predictions $\mathbf{y} = (y_1, ..., y_n)$, 

\section{Experiments}

% We explain our experiment setting in this section.
\subsection{Dataset}
\label{sec:dataset}
We use the SQuAD dataset~\cite{rajpurkar2016squad} to train our models. It is one of the largest general purpose QA datasets derived from Wikipedia with over 100k questions posed by crowdworkers on a set of Wikipedia articles. The answer to each question is a segment of text from the corresponding Wiki passage. The crowdworkers were users of Amazon's Mechanical Turk located in the US or Canada. To obtain high-quality articles, the authors sampled 500 articles from the top 10,000 articles obtained by Nayuki's Wikipedia's internal Page\-Ranks. The question-answer pairs were generated by annotators from a paragraph; and although the dataset is typically used to evaluate reading comprehension, it has also been used in an open domain QA setting~\cite{chen2017reading, wang2018r3}. 
For training/testing answer extraction systems, we pair each paragraph in the dataset with the gold answer spans that it contains. For the question generation system, we pair each sentence that contains an answer span with the corresponding gold question as in \newcite{du2017LearningToAsk}. 

To quantify the effect of using predicted (rather than gold standard) answer spans on question generation (e.g., predicted answer span boundaries can be inaccurate), we also train the models on an augmented ``Training set w/ noisy examples'' (see Table~\ref{tab:qg}). This training set contains all of the original training examples {\em plus} new examples for predicted answer spans (from the top-performing answer extraction model, bottom row of  Table~\ref{tab:tagging}) that {\em overlap} with a gold answer span. We pair the new training sentence (w/ predicted answer span) with the gold question. The added examples comprise 42.21\% of the noisy example training set.
%Proportion of noisy examples that have an overlap with the gold answer is 42.21\%. 

For generation of our one million QA pair corpus, we apply our systems to the 10,000 top-ranking articles of Wikipedia. 
% For each page, only the plain text is extracted and all structured data sections such as lists are stripped.

% For example,~\newcite{chen2017reading} built a two-step system (i.e., search and machine reading), which is able to identify the answer spans given the entire Wikipedia, rather than the associated paragraph to the question.

\subsection{Evaluation Metrics}
For question generation evaluation, we use BLEU~\cite{papineni2002bleu} and METEOR~\cite{2014meteor}.\footnote{We use the evaluation scripts of~\newcite{du2017LearningToAsk}.} BLEU measures average $n$-gram precision vs.\ a set of reference questions and penalizes for overly short sentences. %BLEU-$n$ is the BLEU score that uses up to $n$-grams for counting co-occurrences. 
METEOR is a recall-oriented metric that takes into account synonyms, stemming, and paraphrases.

For answer candidate extraction evaluation, we use precision, recall and F-measure vs.\ the gold standard SQuAD answers. Since answer boundaries are sometimes ambiguous, we compute {\em Binary Overlap} and {\em Proportional Overlap} metrics in addition to {\em Exact Match}. Binary Overlap counts every predicted answer that overlaps with a gold answer span as correct, and Proportional Overlap give partial credit proportional to the amount of overlap~\cite{johansson2010syntactic, irsoy2014opinion}.
% Since the boundaries of answer spans are hard to define, even for human annotators (Wiebe et al., 2005), 
% we use two soft notions of the measures: Binary Overlap counts every overlapping match between a predicted and true expression as correct (Breck et al., 2007; Yang and Cardie, 2012), and Proportional Overlap imparts a partial correctness, proportional to the overlapping amount, to each match (Johansson and Moschitti, 2010; Yang and Cardie, 2012).

% qg performance
\begin{table*}[!htb]
\centering
\resizebox{\textwidth}{!}{%
\begin{tabular}{l|ccc|ccc}
\toprule
\multicolumn{1}{l|}{Models} & \multicolumn{3}{c|}{Training set} & \multicolumn{3}{c}{Training set w/ noisy examples} \\ \midrule
& BLEU-3       & BLEU-4       & METEOR      & BLEU-3                   & BLEU-4                   & METEOR \\ \midrule
Baseline~\cite{du2017LearningToAsk} (w/o answer)  & 17.50        & 12.28        & 16.62       & 15.81                    & 10.78                    & 15.31                   \\
Seq2seq + copy   (w/ answer) & 20.01        & 14.31        & 18.50       & 19.61                    & 13.96                    & 18.19                   \\
\begin{tabular}{@{}l@{}} ContextNQG: Seq2seq + copy  \\ \quad \small (w/ full context + answer) \normalsize \end{tabular}  & 20.31   & 14.58   & 18.84  & 19.57	& 14.05 &	18.19 \\
CorefNQG                                          & \textbf{20.90}        & \textbf{15.16}        & \textbf{19.12}       & \textbf{20.19}                    & \textbf{14.52}                    & 18.59                   \\
\quad - gating                                          & 20.68        & 14.84        & 18.98       & 20.08                    & 14.40                    & \textbf{18.64}                   \\
\quad - mention-pair score                               & 20.56        & 14.75        & 18.85       & 19.73                    & 14.13                    & 18.38    \\
\bottomrule
\end{tabular}}
\vspace{-0.3cm}
\caption{Evaluation results for question generation.}
\vspace{-0.2cm}
\label{tab:qg}
\end{table*}

% ans. extraction
\begin{table*}[!htb]
\small
\centering
\resizebox{\textwidth}{!}{%
\begin{tabular}{l|ccc|ccc|ccc}
\toprule
\multicolumn{1}{l|}{Models} & \multicolumn{3}{c|}{Precision} & \multicolumn{3}{c|}{Recall} & \multicolumn{3}{c}{F-measure}   \\
\midrule
       & Prop.    & Bin.     & Exact   & Prop.   & Bin.    & Exact  & Prop.  & Bin.   & Exact  \\
\midrule
NER  & 24.54 & 25.94 & 12.77 & \textbf{58.20} & \textbf{67.66} & \textbf{38.52} & 34.52 & 37.50 & 19.19 \\
% CNN NER updated            & 24.30 & 25.84 & 12.95 & \textbf{60.35} & \textbf{69.03} & \textbf{39.55} & 34.64 & 37.61 & 19.51 \\
BiLSTM                    & 43.54 & 45.08 & 22.97 & 28.43 & 35.99 & 18.87 & 34.40 & 40.03 & 20.71 \\
BiLSTM w/ NER             & 44.35 & 46.02 & 25.33 & 33.30 & 40.81 & 23.32 & 38.04 & 43.26 & 24.29 \\
BiLSTM-CRF w/ char        & \textbf{49.35} & \textbf{51.92} & \textbf{38.58} & 30.53 & 32.75 & 24.04 & 37.72 & 40.16 & 29.62 \\
BiLSTM-CRF w/ char w/ NER & 45.96 & 51.61 & 33.90 & 41.05 & 43.98 & 28.37 & \textbf{43.37} & \textbf{47.49} & \textbf{30.89}\\
\bottomrule 
\end{tabular}}
\vspace{-0.35cm}
\caption{Evaluation results of answer extraction systems.}
\label{tab:tagging}
\vspace{-0.3cm}
\end{table*}

% \subsection{Training and Implementation Details}
% % We provide our model implementation and training details in this subsection. 
% For the question generation model, the input and output vocabularies are collected from the training data, we keep the 50k most frequent words. The size of word embedding and LSTM hidden states are set to 128 and 256, respectively.We use dropout~\cite{srivastava2014dropout} with probability $p = 0.3$.

% To train our QG model, we first initialize the model parameters randomly using a uniform distribution between $(-0.1, 0.1)$. We use Stochastic Gradient Descent (SGD) as optimization algorithm with a mini-batch size 64. We also apply gradient clipping~\cite{pascanu2013difficulty} with range $(-5, 5)$ during training. The best models are selected based on the perplexity (lowest) on the development set. In all experiments, we use the same split of~\newcite{du2017LearningToAsk} of SQuAD dataset into training, development and test sets.

% We use beam search during decoding to get better results. We set the beam size to 3 in the experiments and corpus generation.

\subsection{Baselines and Ablation Tests}
\label{sec:base}
For question generation, we compare to the state-of-the-art baselines and conduct ablation tests as follows:~{\bf \newcite{du2017LearningToAsk}}'s model is an attention-based RNN sequence-to-sequence neural network (without using the answer location information feature). 
{\bf Seq2seq + copy}\textsubscript{\bf w/ answer} is the attention-based sequence-to-sequence model augmented with a copy mechanism, with answer features concatenated with the word embeddings during encoding. {\bf Seq2seq + copy}\textsubscript{\bf w/ full context + answer} is the same model as the previous one, but we allow access to the full context (i.e., all the preceding sentences and the input sentence itself). We denote it as {\bf ContextNQG} henceforth for simplicity. {\bf CorefNQG} is the coreference-based model proposed in this paper.
{\bf CorefNQG--gating} is an ablation test, the gating network is removed and the coreference position embedding is not refined.
{\bf CorefNQG--mention-pair score} is also an ablation test where all mention-pair $score_i$ are set to zero.

For answer span extraction, we conduct experiments to compare the performance of an off-the-shelf NER system and BiLSTM based systems.
% We also do ablation test to see the effect of different components in our model (refer to Section~\ref{sec:res} for more details).

For {\bf training and implementation details}, please see the Supplementary Material.

\section{Results and Analysis}
\label{sec:res}
\subsection{Automatic Evaluation}
Table~\ref{tab:qg} shows the BLEU-$\{3,4\}$ and METEOR scores of different models. Our CorefNQG outperforms the seq2seq baseline of~\newcite{du2017LearningToAsk} by a large margin. This shows that the copy mechanism, answer features and coreference resolution all aid question generation. In addition, CorefNQG outperforms both Seq2seq+Copy models significantly, whether or not they have access to the full context. This demonstrates that the coreference knowledge encoded with the gating network explicitly helps with the training and generation: it is more difficult for the neural sequence model to learn the coreference knowledge in a latent way. (See input 1 in Figure~\ref{fig:example} for an example.) Building end-to-end models that take into account coreference knowledge in a latent way is an interesting direction to explore. In the ablation tests, the performance drop of CorefNQG--gating shows that the gating network is playing an important role for getting {\em refined} coreference position feature embedding, which helps the model learn the importance of an antecedent. The performance drop of CorefNQG--mention-pair score shows the mention-pair score introduced from the external system~\cite{clark2016improving} helps the neural network better encode coreference knowledge.

% portion
\begin{table}[t]
\centering
\small
\begin{tabular}{l|lll}
\toprule
  & BLEU-3 & BLEU-4 & METEOR \\ \midrule
 \begin{tabular}{@{}l@{}}Seq2seq + copy  \\ \small (w/ ans.) \normalsize \end{tabular}  & 17.81  & 12.30  & 17.11  \\
ContextNQG  & 18.05  & 12.53  & 17.33  \\
CorefNQG  & \textbf{18.46}  & \textbf{12.96}  & \textbf{17.58} \\ \bottomrule
\end{tabular}
%\vspace{-0.35cm}
\caption{Evaluation results for question generation on the portion that requires coreference knowledge (36.42\% examples of the original test set).}
\label{tab:qg-portion}
\vspace{-0.4cm}
\end{table}

To better understand the effect of coreference resolution, we also evaluate our model and the baseline models on just that portion of the test set that requires pronoun resolution (36.42\% of the examples) and show the results in Table~\ref{tab:qg-portion}. The gaps of performance between our model and the baseline models are still significant. Besides, we see that all three systems' performance drop on this partial test set, which demonstrates the hardness of generating questions for the cases that require pronoun resolution (passage context).

We also show in Table~\ref{tab:qg} the results of the QG models trained on the training set augmented with noisy examples with predicted answer spans. There is a consistent but acceptable drop for each model on this new training set, given the inaccuracy of predicted answer spans. We see that CorefNQG still outperforms the baseline models across all metrics.

Figure~\ref{fig:example} provides sample output for input sentences that require contextual coreference knowledge. We see that ContextNQG fails in all cases; our model misses only the third example due to an error introduced by coreference resolution --- the ``city'' and ``it'' are considered coreferent. We can also see that human-generated questions are more natural and varied in form with better paraphrasing.

In Table~\ref{tab:tagging}, we show the evaluation results for different answer extraction models. First we see that all variants of BiLSTM models outperform the off-the-shelf NER system (that proposes all NEs as answer spans), though the NER system has a higher recall. The BiLSTM-CRF that encodes the character-level and NER features for each token performs best in terms of F-measure.
% output examples
\begin{figure}[tb]
\begin{framed}
\small
  \noindent \textbf{Input 1}: The elizabethan navigator, sir francis drake was born in the nearby town of tavistock and was the mayor of plymouth. ... .
%   he was the first englishman to circumnavigate the world and was known by the spanish as el draco meaning "the dragon" after he raided many of their ships. 
\uwave{he died of dysentery in \hl{1596} off the coast of puerto rico}. 
  
  \vspace{0.02cm} \noindent \textbf{Human}: In what year did Sir Francis Drake die ?
  
  \vspace{0.02cm} \noindent \textbf{ContextNQG}: When did he die ?
 
  \vspace{0.02cm} \noindent \textbf{CorefNQG}: When did sir francis drake die ?
  
  \par
  \vspace{0.3cm} 
  \noindent \textbf{Input 2}: american idol is an american singing competition ... . \uwave{it began airing on fox on }\hl{june 11 , 2002}\uwave{, as an addition to the idols format based on the british series pop idol and has since become one of the most successful shows in the history of american television.}
  
  \vspace{0.02cm} \noindent \textbf{Human}: When did american idol first air on tv ?
  
  \vspace{0.02cm} \noindent \textbf{ContextNQG}: When did fox begin airing ?
 
  \vspace{0.02cm} \noindent \textbf{CorefNQG}: When did american idol begin airing ?
  
%   \par
%   \vspace{0.3cm} 
%   \noindent \textbf{Input 2}: ... 
% %   because of its coastal location , the economy of plymouth has traditionally been maritime , in particular the defence sector with over 12,000 people employed and approximately 7,500 in the armed forces . 
%   the plymouth gin distillery has been producing plymouth gin since 1793 , which was exported around the world by the royal navy . \uwave{during }\hl{the 1930s}\uwave{ , it was the most widely distributed gin and has a controlled term of origin.} 
  
%   \vspace{0.02cm} \noindent \textbf{Human}: during what decade was plymouth gin the most widely consumed in the world ?
  
%   \vspace{0.02cm} \noindent \textbf{ContextNQG}: when was it the most distributed term of origin ?
  
%   \vspace{0.02cm} \noindent \textbf{CorefNQG}: when was the plymouth gin distillery the most distributed ?
  
  \par
  \vspace{0.2cm} 
  \noindent \textbf{Input 3}: 
 ... the a38 dual-carriageway runs from east to west across the north of the city . \uwave{within the city it is designated as ` }\hl{the parkway}\uwave{ ' and represents the boundary between the urban parts of the city and the generally more recent suburban areas .}
  
  \vspace{0.02cm} \noindent \textbf{Human}: What is the a38 called inside the city ?
  
  \vspace{0.02cm} \noindent \textbf{ContextNQG}: What is another name for the city ?
  
  \vspace{0.02cm} \noindent \textbf{CorefNQG}: What is the city designated as ?
%   \rule{1\textwidth}{0.5pt}
  
%   \noindent \textbf{Input 4}: The elizabethan navigator, sir francis drake was born in the nearby town of tavistock and was the mayor of plymouth. he was the first englishman to circumnavigate the world and was known by the spanish as el draco meaning "the dragon" after he raided many of their ships. \uwave{He died of dysentery in \hl{1596} off the coast of puerto rico}. 
  
%   \vspace{0.02cm} \noindent \textbf{CorefNQG}: when did sir francis drake die ?

\end{framed}
\vspace{-0.5cm}
\caption{Example questions (with answers highlighted) generated by human annotators (ground truth questions), by our system CorefNQG, and by the Seq2seq+Copy model trained with full context (i.e., ContextNQG).}
% Here \textbf{Human} means ground truth questions.}
\vspace{-0.5cm}
\label{fig:example}
\end{figure}

\vspace{-0.2cm}
\subsection{Human Study}
\begin{table}[t]
% \small
\centering
\resizebox{\columnwidth}{!}{%
    \begin{tabular}{lccccc} 
    \toprule
      & Grammaticality & Making Sense & Answerability & Avg. rank\\\midrule
    % H\&S & 2.83 & 1.91 & 21.88 &2.28 \\
    % Ours & \textbf{3.39} & \textbf{2.97}\textsuperscript{*} &  \textbf{41.75}\textsuperscript{*} & \textbf{1.92}\textsuperscript{**} \\ 
    % Human & 3.71  & 2.61 & 61.96 & 1.53 \\ 
    
% H\&S  & 2.95 & 1.94 & 20.20 & 2.29 \\
% Ours  & \textbf{3.36} & \textbf{3.03}\textsuperscript{*} & \textbf{38.38}\textsuperscript{*} & \textbf{1.94}\textsuperscript{**} \\\midrule
% Human & 3.91 & 2.63 & 66.42 & 1.46 \\

% \begin{tabular}{@{}l@{}} Seq2seq + copy  \\ \small (w/ full context + answer) \normalsize \end{tabular} \\
ContextNQG & 3.793 & 3.836 & 3.892 & 1.768 \\
CorefNQG  & 3.804\textsuperscript{*} & 3.847\textsuperscript{**} & 3.895\textsuperscript{*} & 1.762 \\ \midrule
Human     & \textbf{3.807} & \textbf{3.850} & \textbf{3.902} & \textbf{1.758} \\
    \bottomrule
    % Meaningfulness & 0 & 0 \\ \bottomrule
    \end{tabular}}
  \caption{Human evaluation results for question generation. \small ``Grammaticality'', ``Making Sense'' and ``Answerability'' are rated on a 1--5 scale (5 for the best, see the supplementary materials for a detailed rating scheme), ``Average rank'' is rated on a 1--3 scale (1 for the most preferred, ties are allowed.) Two-tailed t-test results are shown for our method compared to ContextNQG (stat.\ significance is indicated with $^{*}$($p$ < 0.05), $^{**}$($p$ < 0.01).) \normalsize}
%   Two-tailed t-test results are shown for our method compared to H\&S (statistical significance is indicated with $^{*}$($p$ < 0.005), $^{**}$($p$ < 0.001)).}
\label{tab:human}
\vspace{-0.5cm}
\end{table}

We hired four native speakers of English to rate the systems' outputs. Detailed guidelines for the raters are listed in the supplementary materials. The evaluation can also be seen as a measure of the quality of the generated dataset (Section~\ref{sec:dataset}). We randomly sampled 11 passages/paragraphs from the test set; there are in total around 70 question-answer pairs for evaluation. 

We consider three metrics --- ``grammaticality'', ``making sense'' and ``answerability''. The evaluators are asked to first rate the grammatical correctness of the generated question (before being shown the associated input sentence or any other textual context). Next, we ask them to rate the degree to which the question ``makes sense'' given the input sentence (i.e., without considering the correctness of the answer span). Finally, evaluators rate the ``answerability'' of the question given the full context.

Table~\ref{tab:human} shows the results of the human evaluation. Bold indicates top scores. We see that the original human questions are preferred over the two NQG systems' outputs, which is understandable given the examples in Figure~\ref{fig:example}. The human-generated questions make more sense and correspond better with the provided answers, particularly when they require information in the preceding context. How exactly to capture the preceding context so as to \emph{ask} better and more diverse questions is an interesting future direction for research. In terms of grammaticality, however, the neural models do quite well, achieving very close to human performance. In addition, we see that our method (CorefNQG) performs statistically significantly better across all metrics in comparison to the baseline model (ContextNQG), which has access to the entire preceding context in the passage.

% comparing our method (CorefNQG) with the baseline model (ContextNQG), which has the access to the entire preceding context in the passage, our model performs slightly better across all metrics.

\vspace{-0.2cm}
\subsection{The Generated Corpus}
\label{sec:dataset}
\begin{figure}[tb]
\includegraphics[width=6.5cm]{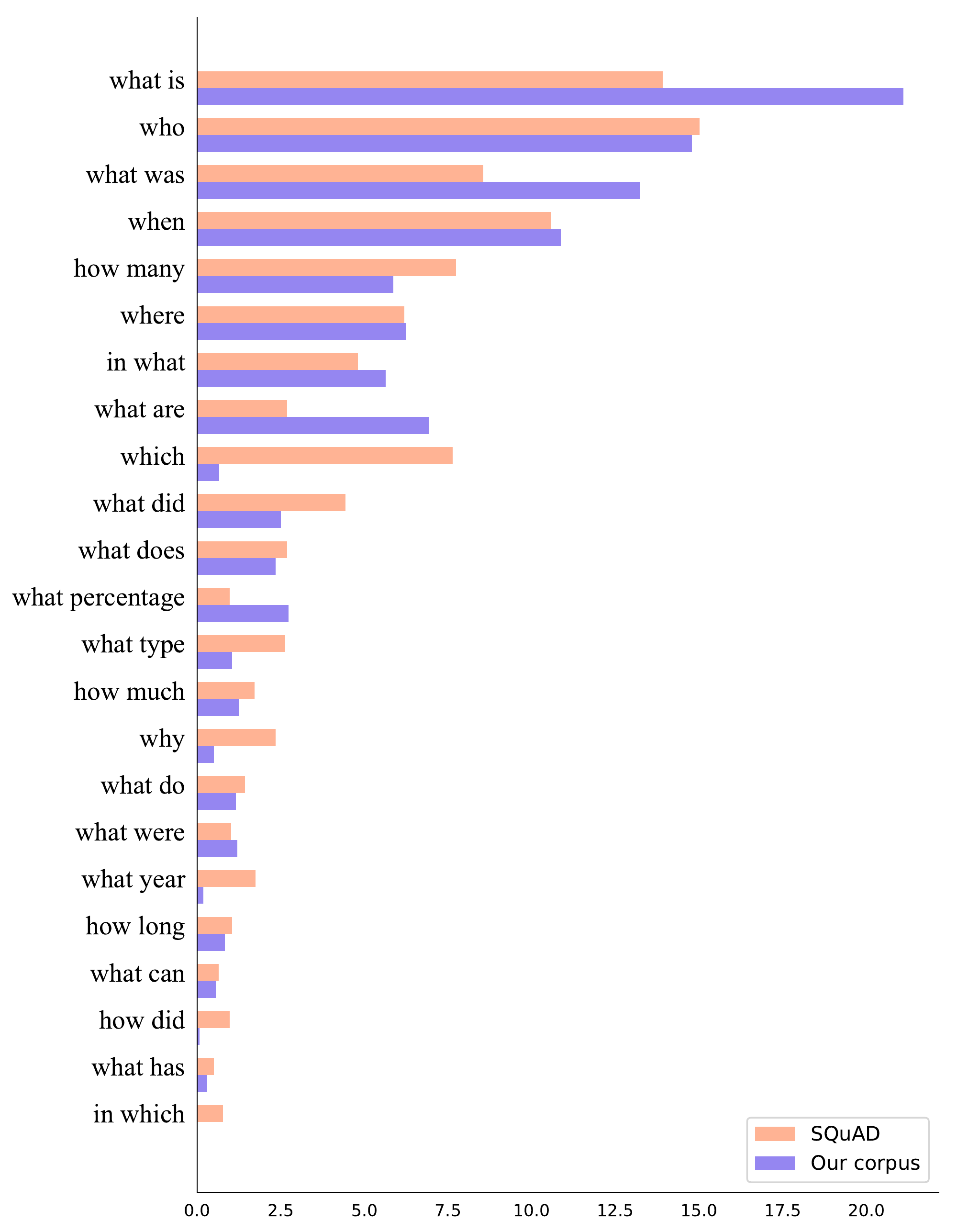}
\vspace{-0.2cm}
\caption{Distribution of question types of our corpus and SQuAD training set. The categories are the ones used in~\newcite{wang2016multi}, we add one more category: ``what percentage''.}
\label{fig:distr}
\vspace{-0.2cm}
\end{figure}

Our system generates in total 1,259,691 question-answer pairs, nearly 126 questions per  article. Figure~\ref{fig:distr} shows the distribution of different types of questions in our dataset vs.\ the SQuAD training set. We see that the distribution for ``In what'', ``When'', ``How long'', ``Who'', ``Where'', ``What does'' and ``What do'' questions in the two datasets is similar. Our system generates more ``What is'', ``What was'' and ``What percentage'' questions, while the proportions of ``What did'', ``Why'' and ``Which'' questions in SQuAD are larger than ours. One possible reason is that the ``Why'',  ``What did'' questions are more \emph{complicated} to ask (sometimes involving world knowledge) and the answer spans are longer phrases of various types that are harder to identify. ``What is'' and ``What was'' questions, on the other hand, are often \emph{safer} for the neural networks systems to ask.
% (e.g., ``what is X?'' is often a valid question, given X is a noun phrase).

% performance on the generated data
\begin{table}[tb]
\centering
\resizebox{\columnwidth}{!}{
\begin{tabular}{l|cccc}
\toprule
& \multicolumn{2}{c}{Exact Match} & \multicolumn{2}{c}{F-1} \\ \midrule
& Dev            & Test            & Dev        & Test       \\ \midrule
% DocReader (w/o pretrain)      &                &                 &            &            \\ \midrule
DocReader~\cite{chen2017reading} &  82.33  &   81.65 &  88.20  &  87.79 \\ \bottomrule
\end{tabular}}
\vspace{-0.2cm}
\caption{Performance of the neural machine reading comprehension model (no initialization with pretrained embeddings) on our generated corpus.}
% \vspace{-0.1cm}
\label{tab:performance}
\end{table}

% examples from corpus
\begin{figure}[!tb]
\begin{framed}
\small
   The United States of America (USA), commonly referred to as the \hl{United States (U.S.) or America}, is a federal republic composed of states, a federal district, \hl{five} major self-governing territories, and various possessions. 
%   Forty-eight of the fifty states and the federal district are contiguous and located in North America between Canada and Mexico.
   ... . The territories are scattered about the \hl{Pacific Ocean and the Caribbean Sea}. Nine time zones are covered. The geography, climate and wildlife of the country are extremely diverse.
  
  \vspace{0.02cm} \noindent \textbf{Q1}: What is another name for the united states of america ?
  
  \vspace{0.02cm} \noindent \textbf{Q2}: How many major territories are in the united states?
 
  \vspace{0.02cm} \noindent \textbf{Q3}: What are the territories scattered about ?
  \vspace{-0.05cm}
\end{framed}
\vspace{-0.4cm}
\caption{Example question-answer pairs from our generated corpus.}
\vspace{-0.4cm}
\label{fig:corpus}
\end{figure}

% question types distr.

In Figure~\ref{fig:corpus}, we show some examples of the generated question-answer pairs. The answer extractor identifies the answer span boundary well and all three questions correspond to their answers. Q2 is valid but not entirely accurate. For more examples, please refer to our supplementary materials.

Table~\ref{tab:performance} shows the performance of a top-performing system for the SQuAD dataset (Document Reader~\cite{chen2017reading}) when applied to the development and test set portions of our generated dataset.   The system was trained on the training set portion of our dataset.  We use the SQuAD evaluation scripts, which calculate exact match (EM) and F-1 scores.\footnote{F-1 measures the average overlap between the predicted answer span and ground truth answer~\cite{rajpurkar2016squad}.} Performance of the neural machine reading model is reasonable. We also train the DocReader on our training set and test the models' performance on the \emph{original} dev set of SQuAD; for this, the performance is around $45.2\%$ on EM and $56.7\%$ on F-1 metric. DocReader trained on the \emph{original} SQuAD training set achieves $69.5\%$ EM, $78.8\%$ F-1 indicating that our dataset is more difficult and/or less natural than the crowd-sourced QA pairs of SQuAD.

% \vspace{-0.1cm}
\section{Conclusion}
% \vspace{-0.1cm}
We propose a new neural network model for better encoding coreference knowledge for paragraph-level question generation. Evaluations with different metrics on the SQuAD machine reading dataset show that our model outperforms state-of-the-art baselines. The ablation study shows the effectiveness of different components in our model. Finally, we apply our question generation framework to produce a corpus of 1.26 million question-answer pairs, which we hope will benefit the QA research community. It would also be interesting to apply our approach to incorporating coreference knowledge to other text generation tasks.

\vspace{-0.1cm}
\section*{Acknowledgments}
We thank the anonymous reviewers and members of Cornell NLP group for helpful comments.

%\bibliography{acl2018}
\bibliography{acl2018}
\bibliographystyle{acl_natbib}

\onecolumn
\appendix

\section{Supplementary Materials}

% Here we provide more sample outputs of our system.
\subsection{Example Question-Answer Pairs from the Corpus}

We provide more examples of QA pairs from the corpus, the red answer spans correspond to the questions in order.

\begin{figure*}[!htb]
\begin{framed}
\footnotesize

\textbf{Paragraph 2} \quad France has long been a global centre of art, science, and philosophy. It hosts Europe's fourth-largest number of cultural UNESCO World Heritage Sites and receives around \hl{83 million} foreign tourists annually, the most of any country in the world. France is a developed country with the world's sixth-largest economy by \hl{nominal GDP} and ninth-largest by purchasing power parity. In terms of aggregate household wealth, it ranks fourth in the world. France performs well in international rankings of education, health care, life expectancy, and human development. France remains a great power in the world, being a founding member of the United Nations, where it serves as one of the five permanent members of the UN Security Council, and a founding and leading member state of the European Union (EU). It is also a member of the Group of 7, \hl{North Atlantic Treaty Organization} (NATO), Organisation for Economic Co-operation and Development (OECD), the World Trade Organization (WTO), and La Francophonie.

\vspace{0.1cm}

\textbf{Questions}: 

\textbf{Q1}: how many foreign tourists does france have ?

\textbf{Q2}: what is france 's sixth-largest economy ?

\textbf{Q3}:  what does nato stand for ?

 \vspace{0.1cm} \hrulefill \vspace{0.2cm}

\textbf{Paragraph 2} \quad The United States embarked on a vigorous expansion across North America throughout the 19th century, displacing American Indian tribes, acquiring new territories, and gradually admitting new states until it spanned the continent by 1848. During the second half of the 19th century, the \hl{American Civil War} led to the end of legal slavery in the country. By the end of that century, the United States extended into the Pacific Ocean, and its economy, driven in large part by the Industrial Revolution, began to soar. The \hl{Spanish-American War} and confirmed the country's status as a global military power. The United States emerged from as a global superpower, the first country to develop nuclear weapons, the only country to use them in warfare, and a permanent member of the United Nations Security Council. It is a founding member of the Organization of American States (UAS) and various other Pan-American and international organisations. The end of the Cold War and the dissolution of the Soviet Union in 1991 left the United States as the world's sole superpower.

% ('Q: in what year did the soviet union join the soviet union ?', u'ans: 1991')
\vspace{0.1cm}

\textbf{Questions}: 

% \textbf{Q1}: in what year did the united states span the continent ?

\textbf{Q1}: what war led to the end of legal slavery ?

\textbf{Q2}:  what was the name of the war that confirmed the country ?

\vspace{0.1cm} \hrulefill \vspace{0.2cm}

\textbf{Paragraph 3} \quad  The \hl{International Standard Name Identifier} (ISNI) is an identifier for uniquely identifying the public identities of contributors to media content such as books, TV programmes, and newspaper articles. Such an identifier consists of \hl{16 digits}. It can optionally be displayed as divided into \hl{four blocks}.

% ('Q: what are some of the media content identifier ?', u'ans: books, TV programmes, and newspaper articles')

\vspace{0.1cm}

\textbf{Questions}: 

\textbf{Q1}: what is an example of a identifier name ?

\textbf{Q2}: how many digits does an identifier have ?

\textbf{Q3}: how long can the identifier be displayed ?

\vspace{0.1cm} \hrulefill \vspace{0.2cm}

\textbf{Paragraph 4} \quad  India, officially the \hl{Republic of India}, is a country in \hl{South Asia}. It is the seventh-largest country by area, the second-most populous country (with over \hl{1.2 billion} people), and the most populous democracy in the world. It is bounded by the Indian Ocean on the south, the Arabian Sea on the southwest, and the Bay of Bengal on the southeast. It shares land borders with Pakistan to the west; China, Nepal, and Bhutan to the northeast; and Myanmar (Burma) and Bangladesh to the east. In the Indian Ocean, India is in the vicinity of \hl{Sri Lanka and the Maldives}. India's Andaman and Nicobar Islands share a maritime border with \hl{Thailand and Indonesia}. Its capital is \hl{New Delhi}; other metropolises include Mumbai, Kolkata, Chennai, Bangalore, Hyderabad and Ahmedabad.

% ('Q: what are some of the media content identifier ?', u'ans: books, TV programmes, and newspaper articles')

\vspace{0.1cm}

\textbf{Questions}: 

\textbf{Q1}: what is india 's country called ?

\textbf{Q2}: where is the republic of india located ?

\textbf{Q3}: how many people are in the second-most populous country ?

\textbf{Q4}: what is the vicinity of india ?

\textbf{Q5}: which two countries did the nicobar islands play a maritime border with ?

\textbf{Q6}: what is the capital of india ?

% \vspace{0.1cm} \hrulefill \vspace{0.2cm}

\end{framed}
% \vspace{-0.5cm}

\label{fig:example}
\end{figure*}

\subsection{Human Rater Guidelines}

We provide the following guidelines for the raters,

\begin{table}[ht]
\centering
\label{my-label}
\begin{tabular}{c|c}
\toprule
Categories     & Rating scheme                                      \\\midrule

\multirow{1}{*}{Grammaticality} & Given only the question itself, is it grammatical?      \\ \midrule

\multirow{1}{*}{Making sense}  & Given just the question and the surrounding context in the passage, \\ 
& does the question make sense?  \\ \midrule

\multirow{2}{*}{Answerability}  & Given just the question and the surrounding context in the passage and the answer \\
& (and regardless of ``Grammaticality'' and ``Making Sense''),\\ 
& is the question answerable by the corresponding answer span?\\ \bottomrule

\end{tabular}
\caption{Guidelines for the raters. For each category, the human raters are required to give a rating ranging from 1 to 5 (5 = fully satisfying the rating scheme, 1 = completely not satisfying the rating scheme, 3 = the borderline cases.}
\end{table}

% \begin{table*}[h]
% \centering
% \begin{tabular}{ c|c|c|c}
% \toprule
% {Score} & \multicolumn{3}{c}{Categories} \\ \hline
%  &  Grammaticality & Making Sense & Answerability \\ \hline 
% & When only looking at the question, is it grammatical?  
% & does the question make sentence \\
% & & (given the passage) 
% % & answerable by the answer span, regardless of the grammaticality and, \\ 
% % is it a good answer? 
% % \hline

% % {5} & a & b & c \\ \hline
% % {4} & a & b & c \\ \hline
% % {3} & a & b & c \\ \hline
% % {2} & a & b & c \\ \hline 
% % {1} & a & b & c \\ 
% \bottomrule

% \end{tabular}
% \vspace{-0.2cm}
% \caption{Guidelines and scores. }
% \vspace{-0.5cm}
% \label{tab:fea_example}
% \end{table*}

\subsection{Training and Implementation Details}
% We provide our model implementation and training details in this subsection. 
For the question generation model, the input and output vocabularies are collected from the training data, we keep the 50k most frequent words. The size of word embedding and LSTM hidden states are set to 128 and 256, respectively. We use dropout~\cite{srivastava2014dropout} with probability $p = 0.3$. The model parameters are initialized randomly using a uniform distribution between $(-0.1, 0.1)$. We use Stochastic Gradient Descent (SGD) as optimization algorithm with a mini-batch size 64. We also apply gradient clipping~\cite{pascanu2013difficulty} with range $(-5, 5)$ during training. The best models are selected based on the perplexity (lowest) on the development set. In all experiments, we use the same split of~\newcite{du2017LearningToAsk} of SQuAD dataset into training, development and test sets.
We use beam search during decoding to get better results. We set the beam size to 3 in the experiments and corpus generation. For the tokenizer used in building the answer extraction system, we use SpaCy.

% % \bibliography{emnlp2017}
% % \bibliographystyle{emnlp_natbib}

% % \bibliography{emnlp2017}
% % \bibliographystyle{emnlp_natbib}

% % \bibliography{emnlp2017}
% % \bibliographystyle{emnlp_natbib}

% % \bibliography{emnlp2017}
% % \bibliographystyle{emnlp_natbib}

\end{document}